# Detection, Retrieval, and Explanation Unified: A Violence Detection System Based on Knowledge Graphs and GAT

Wen-Dong Jiang [a,1], Chih-Yung Chang [a,2] and Diptendu Sinha Roy [c, 3]

[a] *Department of Computer Science and Information Engineering, Tamkang University, New Taipei City 25137, Taiwan.*

[c] *Department of Computer Science and Engineering, National Institute of Technology Meghalaya, Shillong 793003, India.*

*Abstract*—Recently, violence detection systems developed using unified multimodal models have achieved significant success and attracted widespread attention. However, most of them typically suffer from two major issues: lack of interpretability as black-box models and limited functionality, providing only classification or retrieval capabilities. This paper proposes a novel interpretable violence detection system called the T̲hree-i̲n-O̲ne (TIO) System to address these challenges. The proposed TIO system integrates knowledge graphs (KG) and graph attention networks (GAT) to achieve three core functionalities: detection, retrieval, and explanation. Specifically, the system processes each video frame along with text descriptions generated by a large language model (LLM) for a given video containing potential violent behavior. It employs ImageBind to create high-dimensional embeddings for constructing a knowledge graph. Subsequently, the system uses GAT for reasoning, applies lite time series modules to extract video embedding features, and finally connects a classifier and retriever for multi-functional output. Leveraging the interpretability of KG, the system verifies the rationale behind each output during the reasoning process. In addition, this paper proposes several lightweight methods to reduce the resource consumption of the TIO system and improve its efficiency. Extensive experiments conducted on the XD-Violence and UCF-Crime datasets demonstrate the effectiveness of the proposed TIO system. Additionally, the case study reveals an intriguing phenomenon: as the number of bystanders increases, the occurrence of violent behavior tends to decrease.

## 1. Introduction

The primary goal of violence detection is to monitor abnormal events in the real world to prevent violent behaviors and maintain social order [1]. With the rapid advancement of artificial intelligence, researchers are exploring the application of deep learning technologies in surveillance systems [2][3] to replace the inefficiencies and high costs of traditional manual monitoring. In recent years, weakly supervised violence detection (WSVD), also known as weakly supervised video anomaly detection, and violence retrieval (VAR), also referred to as weakly supervised video anomaly retrieval, have become key research directions.

Unlike traditional methods that require frame-level annotations for videos and supervised learning for frame-level training, WSVD achieves frame-level anomaly detection using video-level annotations [4][5]. Meanwhile, VAR focuses on retrieving detailed textual descriptions corresponding to a given

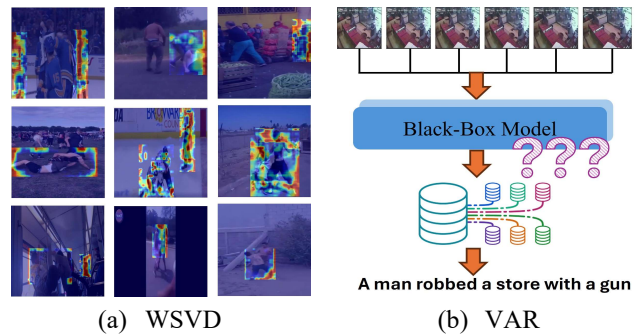

(a)  WSVD    (b)  VAR

**Fig.1**. The interpretable issue in WSVD and VAR task.

video [6], assisting law enforcement in post-event analysis and summarization.[38]-[40]

However, current research on WSVD and VAR tasks [6]-[12] faces two major issues: First, the lack of interpretability in black-box models; second, the need to design separate models for each task. Fig. 1 illustrates the interpretability challenges in WSVD and VAR tasks. In subfigure (a), although the WSVD model successfully identifies violent scenes, interpretability methods like heatmaps reveal that the model focuses on irrelevant areas rather than actual violent scenes. Similarly, in subfigure (b), although the VAR model retrieves the correct textual descriptions, users cannot understand the rationale behind the retrieval results due to the black-box nature of the system.

Moreover, existing research typically designs independent models for WSVD and VAR tasks rather than adopting a unified end-to-end approach, leading to the following problems: (1) Redundant design and training processes increase development and computational costs; (2) Independent models limit information sharing and collaboration, hindering semantic reasoning and performance optimization; (3) Users must separately operate and maintain systems for each task, reducing overall usability and efficiency.

To address these issues, this paper proposes a novel interpretable violence detection system called the Three-in-One (TIO) System. As shown in Fig. 2, unlike previous systems for WSVD and VAR tasks [6]-[12] that suffer from limited functionality and poor interpretability, the TIO system achieves interpretability through a knowledge graph (KG) module and supports end-to-end multi-tasking with a dual-branch design. Specifically, given video data with violent behaviors and corresponding labels, the system first uses an LLM to generate

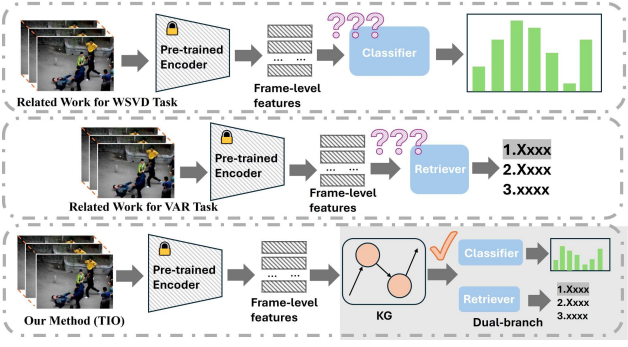

**Fig.2**. Comparisons of different paradigms for WSVD and VAR.

keywords related to the labels. These, along with video data are passed through a frozen ImageBind encoder [13] to produce unified multimodal embeddings. Next, a knowledge graph is constructed and refined through a graph attention network (GAT) [14]. Temporal sequence analysis is then performed using a lightweight temporal module, followed by a classifier and retriever for end-to-end multi-task functionality. During usage, users not only receive system decisions and retrieval results but also gain explanations through KG-based triplet relationships (entity1-relation-entity2).

In practice, triplet construction is defined as follows: entity1 represents high-dimensional video embeddings, relations are LLM-generated keyword embeddings, and entity2 comprises object information extracted using the pre-trained Yolo-World [15], supplemented with semantic information via Concept Net [20]. Lightweight design focuses on GAT and temporal modules. For the GAT module, based on Kernel function adjust Euclidean distance is used instead of inner product operations, simplifying system complexity. For the temporal module, distance-based computation replaces traditional similarity calculations, improving computational efficiency and model performance. Detailed implementation is described in **Section 4**.

Experiments on benchmark datasets XD-Violence [18] and UCF-Crime [19] demonstrate the feasibility of the proposed TIO system. In summary, the main contributions of this paper are as follows:

1. **Incorporation of a Knowledge Graph (KG) module and dual-branch design:** The proposed system achieves end-to-end processing for weakly supervised violence detection (WSVD) and violence retrieval (VAR) tasks, significantly enhancing the interpretability of the model, making system decisions and retrieval results more transparent.
2. **Innovative use of multimodal unified embeddings and Euclidean distance for model optimization**: Unlike previous methods, the proposed TIO system employs a frozen ImageBind encoder to generate unified multimodal embeddings and uses Euclidean distance instead of inner product operations in the Graph Attention Network (GAT) of the knowledge graph. This approach substantially reduces system complexity. Additionally, the temporal module leverages distance-based metrics rather than traditional similarity calculations, improving computational efficiency and model performance.
3. **Unified multi-task model design for enhanced system efficiency and user experience**: The proposed TIO system replaces the traditional approach of designing separate models for WSVD and VAR tasks with a unified multi-task model. This reduces redundancy in model design and training, facilitates information sharing and collaborative use, minimizes user operation and maintenance efforts, and achieves dual improvements in semantic reasoning and performance optimization.

The remainder of the paper is organized as follows. **Section 2** discusses and compares previous relevant studies. **Section 3** describes the Assumptions and problem formulation in detail. **Section 4** details the proposed TIO system in this paper. **Section 5** provides experiments and performance evaluation. The conclusion is discussed in **Section 6**.

## 2. Related Work

This section reviews the related literature, divided into two parts: Weakly Supervised Violence Detection, and Violence Retrieval.

### 2.1. Weakly Supervised Violence Detection

*Sultani et al.* [19] and *Hasan et al.* [20] were the first to introduce the Multi-Instance Learning (MIL) model into Supervised Violence Monitoring (SVM). Their approach packaged surveillance videos and fed them into I3D and C3D networks for binary classification of violent and non-violent events. *Ji and Lee* [21] proposed the One-Class Support Vector Machine (OCSVM) model for violence detection. However, due to the limitations of 3D network-based designs in terms of real-time performance and hardware requirements in practical applications, research gradually shifted toward combining video frame extraction and temporal sequence modules for violence detection. Subsequent studies focused on utilizing self-attention mechanisms, Transformer, or Graph ConvNet (GCN) to capture temporal and contextual relationships within video content.

*Zhong et al.* [22] proposed a GCN-based method to calculate feature similarity and temporal consistency between video segments for monitoring violent behavior. *Tian et al.* [23] proposed Robust Temporal Feature Magnitude (RTFM) which utilized a self-attention network to capture the global temporal context of violent behaviors in violent videos. *Wu et al.* [24] introduced S3R, which models anomalies at the feature level by exploring the synergy between dictionary-based representation and self-supervised learning. *Li et al.* [25] proposed Spatial-Temporal Relation Learning (STRL), which jointly utilizes spatial and temporal evolution patterns for feature learning.

It is noteworthy that in certain practical scenarios, only single-modal cameras are typically available, and traditional multi-modal designs are often inapplicable. Thanks to pre-trained models such as CLIP [26], which align images and text to accomplish multi-modal tasks, and Imagebind [13], which unified data from images and seven different heterogeneous modalities, it became possible to achieve multi-modal tasks even in single-modality scenarios. Some studies attempted to deploy these pre-trained models for violence detection. For instance, *Wang et al.* [27] proposed ActionClip, which enhances video representation with richer semantic language supervision and supports zero-shot action recognition without additional labeled data or parameter requirements. *Zanella et al.* [28] combined large language and vision (LLV) models (e.g.,

CLIP) with multi-instance learning for joint violence detection. *Wu et al.* [7] introduced Spatio-Temporal Prompting (STPrompt), a CLIP-based three-branch architecture to address classification and localization in violence detection. *Wu et al.* [8] proposed Video Anomaly CLIP (VadCLIP), a simple yet powerful baseline that efficiently adapted pre-trained image-based visual-language models for robust general video understanding. In 2025, *Sanggeon et al.* [30] proposed MissionGNN, which integrated Imagebind with LLM to construct a multi-level Graph Neural Network for anomaly detection.

Although these methods have made significant progress in Weakly Supervised Violence Detection, the models designed in these studies are essentially black boxes, raising serious concerns about interpretability. Furthermore, some methods, such as GCN, GNN, or Transformer-based networks for temporal sequence analysis, inherently demand significant computational resources.

## 2.2. Violence Retrieval

Events in videos typically reflected the interactions between actions and entities evolving over time, and people tended to use retrieval methods to obtain comprehensive descriptions. Benefiting from the success of cross-modal pre-trained models, some studies focused on audio-text and audio-video retrieval. *Liu et al.* [31] proposed Collaborative Experts (CE), a collaborative expert model that compressed multimodal information into compact representations. It utilized pre-trained semantic embeddings along with specific cues such as Automatic Speech Recognition (ASR) and Optical Character Recognition (OCR) to query videos through text. *Gabeur et al.* [32] proposed the Multi-Modal Transformer (MMT), a multimodal transformer model designed to jointly encode different modalities in videos, enabling inter-modal attention.

*Wang et al.* [33] introduced Text-to-Video Vector of Locally Aggregated Descriptors (T2VLAD), which implemented an efficient global-local alignment method. It adaptively aggregated multimodal video sequences and textual features through a shared set of semantic centers and computed local cross-modal similarities between video and textual features within the same centers. *Ma et al.* [34] proposed Cross-Modal Clip (XClip), which calculated the correlation between coarse-grained features and each fine-grained feature through coarse-to-fine contrast. It filtered out unnecessary fine-grained features guided by coarse-grained features during similarity computation, thereby improving the accuracy of video-to-text retrieval. *Lei et al.* [35] introduced Cross-Modal Late Fusion (XML), which adopted a late fusion design and a novel convolutional start-end detector (Convolutional Start-End Detector, ConvSE) to achieve efficient retrieval. *Wu et al.* [6] proposed the Anomaly-Led Attention Network (ALAN), which developed an anomaly-led sampling method to enhance the fine-grained semantic association between violent videos and text, further matching cross-modal content using two complementary alignment strategies.

Although these methods achieved significant success in retrieval tasks, these models still lacked interpretability. During inference, it was difficult for users to understand the reasons behind successful matches. Additionally, the models designed for these tasks were typically limited in functionality, focusing solely on retrieval while being unable to classify violent content in a timely manner.

## 3. Assumptions and Problem formulation

This section presents the assumptions and problem statements of this study, divided into two parts: Detection and Retrieval.

### 3.1. Detection

Given a monitoring video sequence $\mathcal{V}$ of duration $\tau > 0$, partitioned into $N$ equal-length temporal segments $\{\Phi_l\}_{l=1}^N$. The principal aim is to detect anomalies (violent events). Let $\mathcal{V} = \{\Phi_t\}_{t=1}^N$ where each $\Phi_l \subseteq \mathcal{V}$ may consist of non-violent $(\widehat{\mathcal{V}_l})$ and violent $(\mathcal{V}_l)$ portions, so $\Phi_l = \widehat{\mathcal{V}_l} \cup \mathcal{V}_l$.

Let $\mathcal{C} = \{c_k\}_{k=1}^K \cup \{\hat{c}\}$ denote all classes, with $c_k$ denote the $k$-th violence class and $\hat{c}$ represent the non-violence class, and let $\mathcal{L} = \{\ell_k\}_{k=1}^K \cup \{\hat{\ell}\}$ be the corresponding labels. Assume each $\Phi_l$ containing violence belongs to some $c_k$, where $k \in \{1,...,K\}$.

Consider a detection mechanism $\mathcal{M} \in \mathfrak{M}$, which outputs for each $\Phi_l$ a score $\mathcal{P}_l \in [0,1]$. Let $\theta \in [0,1]$ be a decision threshold inducing $\delta_l^M(\theta) = \mathbb{1}[\mathcal{P}_l \geq \theta]$. Let $\delta_l = \mathbb{1}[\mathcal{V}_l \neq \emptyset]$ be the ground truth.

Let $TP_l, TN_l, FP_l$ and $FN_l$ represent True Positive, True Negative, False Negative and False Positive respectively, of the prediction result of input $\widehat{\mathcal{V}_l}$ by applying mechanism $\mathcal{M}$. Then $TP_l = \delta_l \delta_l^M(\theta)$, $TN_l = (1-\delta_l)(1-\delta_l^M(\theta))$, $FP_l = (1-\delta_l)\delta_l^M(\theta)$ and $FN_l = \delta_l(1-\delta_l^M(\theta))$, respectively.

Let $TP, TN, FP$ and $FN$ represent the cumulative True Positive, True Negative, False Positive, and False Negative, respectively. Summations over $l = 1,...,N$ yield $TP = \sum TP_l$, $TN = \sum TN_l$, $FP = \sum FP_l$ and $FN = \sum FN_l$, respectively.

Let $\mathcal{P}^{\mathcal{M}}, \mathcal{R}^{\mathcal{M}}$ denote the Precision and Recall of the predictions by applying mechanism $\mathcal{M}$ to predict a given video $\mathcal{V}$. The values of $\mathcal{P}^{\mathcal{M}}, \mathcal{R}^{\mathcal{M}}$ can be calculated by $\mathcal{P}^{\mathcal{M}} = \frac{TP}{TP+FP}$ and $\mathcal{R}^{\mathcal{M}} = \frac{TP}{TP+FN}$, respectively.

Let $\{\theta_j\}_{j=1}^J \subset [0,1]$ denote a finite collection of thresholds. Let $\mathcal{AP}^{\mathcal{M}}$ denote the Average Precision of mechanism $\mathcal{M}$. The $\mathcal{AP}^{\mathcal{M}}$ can be calculated by Exp. (1):

$$\mathcal{AP}^{\mathcal{M}} = \sum_{j=1}^{J-1} \left(\mathcal{R}^{\mathcal{M}}(\theta_{j+1}) - \mathcal{R}^{\mathcal{M}}(\theta_j)\right) \times \mathcal{P}^{\mathcal{M}}(\theta_{j+1}).$$

Similar to [1]-[5] [7][13][19]-[28], the first objective of violence detection in this paper is to develop mechanism $\mathcal{M}^{best}$ that satisfies $\mathcal{M}^{best} = arg \max_{\mathcal{M} \in \mathfrak{M}}(\mathcal{AP}^{\mathcal{M}})$.

Let $\text{TPR}(\theta) = \frac{TP(\theta)}{TP(\theta)+TN(\theta)}$ and $\text{FPR}(\theta) = \frac{FP(\theta)}{TP(\theta)+TN(\theta)}$ denote the true-positive and false-positive rates as functions of $\theta$, respectively. Let $\mathcal{A}_{\mathcal{M}}$ denote the $AUC$ of mechanism $\mathcal{M}$. The value of $\mathcal{A}_{\mathcal{M}}$ can be calculated by Exp. (2):

$$\mathcal{A}_{\mathcal{M}} = \int_0^1 TPR(\theta)\,d(FPR(\theta)) = \int_0^1 \frac{TP(\theta)}{TP(\theta)+FN(\theta)} d\left(\frac{FP(\theta)}{FP(\theta)+TN(\theta)}\right).$$

Similar to [1]-[5] [7][13][19]-[28], the second objective of violence detection in this paper is to develop mechanism $\mathcal{M}^{best}$ that satisfies $\mathcal{M}^{best} = arg \max_{\mathcal{M} \in \mathfrak{M}}(\mathcal{A}_{\mathcal{M}})$.

## 3.2. Retrieval

Given a scenario involving both time-series data and multiple triplet relationships, let $\{\alpha_t\}_{t=1}^{T}$ represent a sequence of observations recorded at each time $t$, and let $\{(h_n, r_n, t_n)\}_{n=1}^{N}$ denote a collection of triplets, where $h_n, r_n$, and $t_n$ are the head entity, relation, and tail entity, respectively.

By applying a time-series module, temporal context can be integrated with the triplet semantics, resulting in a set of embeddings $\{z_n\}_{n=1}^{N} \subseteq \mathbb{R}^d$. Each embedding $z_n$ corresponds to a triplet $(h_n, r_n, t_n)$ as well as its associated temporal information $\{\alpha_t\}$.

Consider a query vector $q \in \mathbb{R}^d$. Let $s(z_n, q) \in \mathbb{R}$ denote the similarity function, which measures the relevance of $z_n$ to $q$.

Based on the similarity scores, all embeddings $z_n$ are ranked in descending order of $s(z_n, q)$. Denote the corresponding ranking function as $\pi: \{1, \ldots, N\} \to \{1, \ldots, N\}$, such that: $s(z_{\pi(1)}, q) \geq s(z_{\pi(2)}, q) \geq \cdots \geq s(z_{\pi(N)}, q)$.

Let $\delta_n \in \{0,1\}$ denote an indicator function that specifies whether $z_n$ is ground-truth relevant to the query $q$. Let $\sum_{n=1}^{N} \delta_n$ represent the total number of relevant embeddings. Define R@k as the fraction of relevant items retrieved within the top-$k$ positions out of all relevant items. For any $k \in \mathbb{N}$, it is defined as Exp. (3):

$$R@k = \sum_{r=1}^{k} \delta_{n(r)} \Big/ \sum_{r=1}^{k} \delta_n.$$

When $k \in \{1, 5, 10\}$, this corresponds to R@1, R@5, R@10, respectively. Similar to previous works [6] [31]-[35], this study aims to maximize retrieval performance. Within the retrieval strategy space $\mathfrak{R}$, the objective is to identify the optimal retrieval strategy $\mathfrak{R}^{best}$ such that: $\mathfrak{R}^{best} = \arg\max_{\mathcal{R} \in \mathfrak{R}} \{R@1, R@5, R@10\}$.

The above content constitutes the Assumptions and Problem Formulation of this paper, The next section will introduce the proposed TIO system in detail.

## 4. The proposed TIO system

This section provides a detailed introduction to the proposed TIO system. Unlike previous systems designed specifically for violence detection [7], [19]-[28] or violence retrieval [6], [31]-[35], the proposed TIO system in this paper offers two significant advantages: first, TIO is a multifunctional system capable of both violence detection and retrieval tasks; second, the design of TIO incorporates interpretability.

Figure 4 illustrates the operational framework of the proposed TIO system. The system is divided into three main components: feature extraction, lightweight temporal sequence analysis and classification, and retrieval task deployment.

Specifically, in terms of feature extraction, the TIO system leverages LLM and ImageBind pre-trained modules to process the information in each frame of training data, constructing a knowledge graph and optimizing it using a GAT module. After the feature extraction, the compressed representation of each video frame is further processed by a lightweight temporal sequence module to generate a high-dimensional embedding for the entire video. Finally, the system connects to both a classifier and a retriever to achieve its multitasking objectives. During inference, through the output of the triplet relationships in the knowledge graph, users can clearly understand the basis of the model's decisions, thereby achieving interpretability.

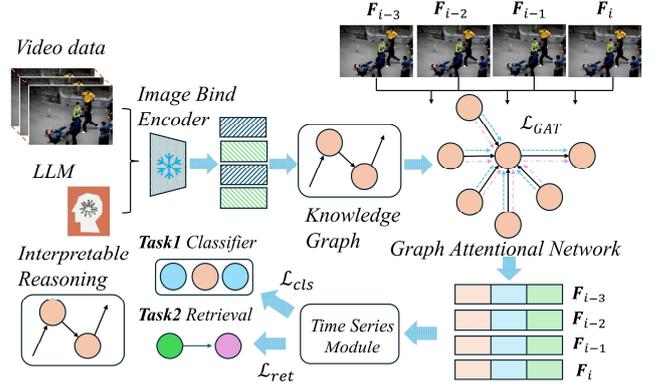

**Fig.4.** The proposed TIO System

The specific designs of these three components are detailed in the following subsections: **Section 4.1** introduces the construction of the knowledge graph and GAT optimization, **Section 4.2** describes the design of the lightweight temporal sequence module, and **Section 4.3** explains how to integrate the classifier and retriever to complete the system training.

### 4.1. Feature Extraction

This section provides a detailed explanation of the feature extraction functionality of the TIO system. Notably, unlike previous works [6][7][19][28][31]-[35], which directly employed pre-trained modules for feature extraction, this study adopts a novel approach by constructing a knowledge graph and leveraging GAT-based reasoning for frame-level feature extraction. This method not only captures the semantic relationships between multimodal data but also enhances the interpretability and discriminative power of features through structured reasoning, thereby improving the accuracy of recognition and retrieval.

The feature extraction process is divided into two steps: the construction of the knowledge graph and the refinement via GAT. The specific details are described as follows:

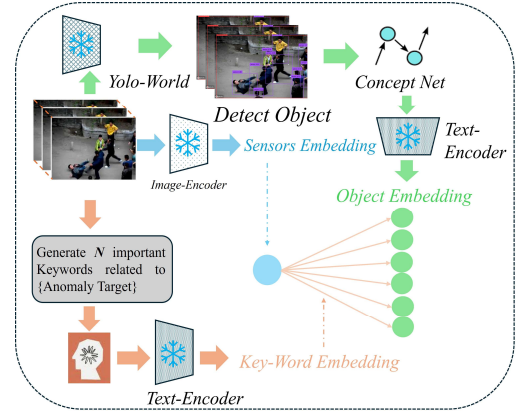

**Fig.5.** The details of KG

First, the construction of the knowledge graph: Fig. 5 illustrates the overall framework for generating the task-specific knowledge graph (KG). Specifically, let $\mathfrak{I} = \{I_t\}_{t=1}^{T}$ denote video sequence, where each frame $I_t$ is drawn from an underlying space $\Omega_{image} \subseteq \mathbb{R}^d$ to map each frame $I_t$ to its corresponding $d$-dimensional embedding $v_t^{image} = E_I(I_t)$, thereby yielding the set of frame-level entities $V_{image} = \{v_1^{image}, v_2^{image}, \ldots, v_T^{image}\} \subseteq \mathbb{R}^d$. Next, the predefined label set $L = \{l_1, l_2, \ldots, l_n\}$ and invoke a large language model $f_{LLM}(.)$ generate a collection of keyword-based relation types $K = \{k_1, k_2, \ldots, k_m\} \subseteq \mathcal{R}$, where each $k_j$ is derived from some $l_i$ via $k_j = f_{LLM}(l_i)$. Subsequently, for each frame $I_t$, the pretrained YOLO-World [15] obtains a set of detected objects $\mathcal{O}_t = \{o_{t,1}, o_{t,2}, \ldots, o_{t,n}\}$, where each $o_{t,i}$ is represented as $(class_{t,i}, bbox_{t,i}) \in \Omega_{class} \times \Delta$, with $\Omega_{class}$ denoting the space of possible object classes and $\Delta \subseteq \mathbb{R}^4$ denoting the space of bounding box coordinates. For each detected object $o_{t,j}$, the external textual description $d_{t,i}$ from ConceptNet [20] and then feed $d_{t,i}$ into a pretrained text encoder $E_T: \Omega_{text} \to \mathbb{R}^d$, where $\Omega_{text}$ is the textual input space to obtain the semantic embedding $v_{t,i}^{object} = E_T(d_{t,i})$, thereby forming the object-level entity set $V_{object} = \bigcup_{t=1}^{T}\{v_{t,i}^{object} | i = 1, \ldots, n_t\} \subseteq \mathbb{R}^d$. Define a multimodal triple $\tau \in V_{image} \times K \times V_{object}$ of the form $\tau = (v_t^{image}, k_j, v_{t,i}^{object})$ to represent the semantic relation $k_j$ between the frame-level embedding $v_t^{image}$ and the object-level embedding $v_t^{object}$.

By aggregating all such possible triples across every $t \in \{1, \ldots, T\}$, every $i \in \{1, \ldots, n_t\}$, and every $j \in \{1, \ldots, m\}$, the edge set $E$ represents in:

$$E = \bigcup_{t=1}^{T} \bigcup_{i=1}^{n_t} \bigcup_{j=1}^{m}\{(v_t^{image}, k_j, v_{t,i}^{object})\} \subseteq V_{image} \times K \times V_{object}.$$

Finally, the node set $V = V_{image} \cup V_{object}$ and the edge set $E$ into a directed KG $G = (V, E)$. One may further encode the relationship structure of $G$ in a tensor form $A \in \{0,1\}^{|V| \times |V| \times |K|}$ by letting $A_{(v,v,j)} = 1$ if and only if $(u, k_j, v) \in E$ for some $u \in V_{image}, u \in V_{object}$, and $A_{(v,v,j)} = 0$ otherwise, enabling a higher-order adjacency tensor representation that captures not only the links between entities but also the specific type of relation assigned by each $k_j$.

After constructing the knowledge graph, the GAT module was adjusted. Unlike previous studies [22][23] that primarily utilized GNN architectures for time-series analysis, this study innovatively applied GAT to refine the knowledge graph.

The main advantage of this approach lies in the attention mechanism of GAT, which effectively assigns weights to the nodes and edges in the knowledge graph. This allows the refinement process to more precisely capture key semantic relationships within the graph. As a result, the structural integrity of the knowledge graph is enhanced, significantly improving its performance.

An intriguing phenomenon was observed during experimentation. The existing unified cross-modal pre-trained model, ImageBind [13], has a high dimensionality, leading to a high computational cost for similarity comparisons based on attention mechanisms. Furthermore, for violent events of the same category and their corresponding textual descriptions, the feature vectors extracted by ImageBind exhibit high similarity.

To address these issues, this study proposes a Euclidean distance-based correction method using an exponential kernel function as an alternative to the computationally expensive similarity comparisons and dot product operations.

Specifically, after constructing the task-related knowledge graph $G = (V, E)$ with each node $v \in V$ associated with a feature vector $h_v \in \mathbb{R}^d$. Let $d_{u,v}$ denote the pairwise Euclidean distance. First, calculates the $d_{u,v}$ for any edge $(u, k_j, v) \in E$ as shown in Exp. (4):

$$d_{u,v} = \sum_{k=1}^{d}|h_{u,k} - h_{v,k}|^2 = \|h_u - h_v\|_2^2.$$

Then it finds $d_{max} = \max_{(u,v) \in E} d_{u,v}$ and $d_{min} = \min_{(u,v) \in E} d_{u,v}$, applies interval normalization in Exp. (5):

$$d'_{uv} = (d_{u,v} - d_{min})/(d_{max} - d_{min}),$$

and use a Gaussian kernel to obtain the edge weight $w_{u,v}$ in Exp. (6):

$$w_{u,v} = exp(-(d'_{uv})^2 / \sigma^2).$$

Rather than introducing any feature-based similarity or inner-product term, one can define the attention score purely $e_{u,v} = w_{u,v}$. The attention score applies to a *softmax* over neighbors $N(u)$ to obtain in Exp. (7):

$$\alpha_{u,v} = exp(e_{u,v})/\Sigma_{k \in N(u)} exp(e_{u,k}),$$

and finally update the node feature via $h'_u = \sigma(\sum_{v \in N(u)} \alpha_{u,v} h_v)$, so that attention depends solely on the geometrical separation of nodes as captured by the distance-based kernel. Proof1 demonstrates that the proposed method is lightweight compared to the multi-head dot product approach.

**Proof1**: There are $n$ high-dimensional vectors $\{x_1, \ldots, x_n\} \subset \mathbb{R}^D$. The goal is to calculate the "similarity/attention scores" for all pairwise combinations $(x_i, x_j)$, where $i, j = 1, \ldots, n$. Two methods are compared in terms of computational cost: Multi-head Dot Product Method and Distance Kernel Method.

Let $H$ denote the Number of heads. Let $W_Q, W_k \in \mathbb{R}^{d \times D}$ denote the Projection dimension for each head. For each vector $x_i$, the following are computed $q_i^{(h)} = W_Q^{(h)} x_i$ and $k_i^{(h)} = W_K^{(h)} x_i$, where $W_Q^{(h)}, W_K^{(h)} \in \mathbb{R}^{d \times D}$. Each transformation requires $O(Dd)$ time, with 2 transformations ($Q$ and $K$) per head. For $n$ vectors and $H$ heads, the total cost is $T_{proj} = H \cdot n \cdot 2 \cdot O(Dd) = O(HnDd)$.

For all pairwise combinations $(i, j)$, the dot product is computed $q_i^{(h)\top} k_j^{(h)}$, where each dot product requires $O(d)$ time. The number of pairwise combinations is approximately $n^2 / 2$, and with $H$ heads, the total cost is $T_{atten} = H \cdot O(n^2) \cdot O(d) = O(Hn^2 d)$. The overall complexity of the multi-head method is $T_{multi-head} = O(HnDd) + O(Hn^2 d)$.

For each pair $(x_i, x_j)$, the squared Euclidean distance is calculated in Exp. (4), which requires $O(D)$ time. For approximately $n^2 / 2$ pairs, the total cost is $T_{dist-calc} = O(n^2 D)$.

For each pairwise distance $d^2(x_i, x_j)$, the following kernel is applied in Exp. (5), which is a scalar operation with cost $O(1)$.

Thus, the total cost remains $T_{kernel} = O(n^2D)$. Sufficient conditions for $T_{kernel} < T_{multi-hea}$. In the following conditions hold $n^2D < C_1 \cdot HnDd$ and $n^2D < C_2 \cdot Hn^2d$, where $C_1$, $C_2$ are constants, then $n^2D < HnDd + Hn^2d \Rightarrow T_{kernel} < T_{multi-hea}$.

In summary, by constructing the aforementioned process, each video frame can obtain a high-dimensional vector representation at the frame level $\{h_v\}_{v \in V_{image}}$ through GAT during the input stage. Subsequently, these vectors are further processed using a lightweight temporal sequence module.

### 4.2. Lightweight Temporal Sequence Module

Till now, after completing frame-level feature extraction, the extracted features capture instantaneous information but lack the global temporal context critical to video understanding. In this section, a lightweight temporal sequence module is proposed, analogous to the standard Receptance Weighted Key Value (RWKV) [36] encoder, incorporating self-attention, Layer Normalization (LN), and a Feed-Forward Network (FFN). Since the positions of frames in sequential data are already fixed, positional encoding is unnecessary when training the temporal sequence module.

Compared to prior works [22]-[27], the primary distinction lies in the computation method of self-attention. This method is based on relative distances rather than feature similarity, making the model lightweight. By focusing solely on the relative distances between frames and avoiding the computation of high-dimensional feature similarity, this approach effectively reduces the complexity of attention computation, thereby lowering computational costs and improving training and inference efficiency.

Specifically, let $V = \{v_1, v_2, \ldots, v_n\}$ denote the set of frames in the video. Using the GAT module, the embeddings $h_{vi} \in \mathbb{R}^d$ are obtained. All frame embeddings are integrated into a matrix $H = [h_{v1}, h_{v2}, \ldots, h_{vN}] \in \mathbb{R}^{N \times d}$.

Define $A_t \in \mathbb{R}^{N \times N}$, where $N$ represents the total number of frames. For any $1 \leq i, j \leq N$, its elements are defined in Exp. (8):

$$(A_t)_{i,j} = exp(-|i-j|/\sigma),$$

where $\sigma > 0$ is a hyperparameter that controls the degree of temporal distance decay. This design ensures that $A_t$ depends solely on the relative distances between frame indices, reducing reliance on high-dimensional feature similarity.

To enhance numerical stability, a symmetric normalization strategy is adopted. Define the diagonal degree matrix $D \in \mathbb{R}^{N \times N}$, with diagonal elements given by Exp. (9):

$$(D)_{i,i} = \sum_{j=1}^{N}(A_t)_{i,j},$$

and construct the symmetrically normalized adjacency matrix $\tilde{A}_t = D^{-1/2}A_tD^{-1/2}$. This symmetric normalization ensures the stability of gradient propagation and numerical operations during feature fusion.

In practice, the proposed temporal sequence module consists of two stages to integrate and enhance temporal context in the embedding matrix $H$.

The first stage is temporal context fusion. The normalized temporal adjacent matrix $\tilde{A}_t$ is used to weight the frame embedding matrix $H$, as shown in Exp. (10):

$$H' = softmax(\tilde{A}_t)H \in \mathbb{R}^{N \times d},$$

where $softmax(\tilde{A}_t)$ performs *softmax* normalization row-wise on $\tilde{A}_t$ ensuring a reasonable distribution of weighting coefficients. Subsequently, Layer Normalization (LN) is applied to $H'$ to stabilize feature distributions, as shown in Exp. (11):

$$H'' = LN(H').$$

The next stage involves non-linear transformation and residual connection. The fused features $H''$ are subjected to a FFN for non-linear transformation, with residual connections preserving the original information. The operation is as follows in Exp. (12):

$$\begin{aligned}H''' &= LN(FFN(H'') + H'') \\ &= LN(ReLU(xW_1 + b_1)W_2 + b_2),\end{aligned}$$

where $W_1$, $W_2$ are the parameter matrices for linear transformations, $b_1$, $b_2$ are bias vectors, and the $ReLU$ function provides non-linear mapping capabilities, enhancing the representational power of the model.

### 4.3. Classifier and Retriever

After completing the time-series computation, the classifier and retriever are integrated. The classifier is used to determine which predefined category the current image frame (or frame-level feature) belongs to. The retriever is used to locate the corresponding textual description in the database, enabling cross-modal retrieval.

Specifically, for the classifier, the loss function is defined as $\mathcal{L}_{cls}$, as shown in Exp. (13):

$$\underbrace{\mathcal{L}_{cls} = -\frac{1}{N}\sum_{i=1}^{N}\sum_{c=1}^{C}\delta_{y_i,c}\log p_{i,c}}_{For\ Classification},$$

Where $N$ is the number of samples or video frames in a batch, $C$ is the total number of categories, and $\delta_{y_i,c}$ is the indicator function defined as: $\delta_{y_i,c} = \begin{cases}1, if\ y_i = c \\ 0, if\ y_i \neq c\end{cases}$. The $p_{i,c} = P(y_i = c|z_i)$ represents the predicted probability by the classifier that the $i$-th sample belongs to category $c$ and $z_i$ is the feature representation of the frame or sample.

For the retriever, the loss function is defined as $\mathcal{L}_{ret}$, as shown in Exp. (14):

$$\underbrace{\mathcal{L}_{ret} = \sum_{(i,j)\in\Omega}max\{0, \alpha + D(q_i, t_j^+) - (q_i, t_j^-)\}}_{For\ Retrivel},$$

Where $\Omega$ is the set of indices for all text-video pairs; $q_i$ represents the embedding of the $i$-th image; $t_j^+$ and $t_j^-$ represent the text vectors matching $q_i$ (positive sample) and not matching $q_i$ (negative sample), respectively; $D(q_i, t_j^+)$ is the distance function; and $\alpha > 0$ is the margin hyperparameter. This loss encourages the distance of correctly matched image-text pairs to be smaller than that of mismatched pairs, enabling effective retrieval judgments.

Through backpropagation with these two loss functions, parameters in the GAT module are also updated. A self-regularization loss is designed for the GAT module, expressed as $\mathcal{L}_{gat}$ as shown in Exp. (15):

$$\underbrace{\mathcal{L}_{GAT} = \lambda \sum_{(i,j) \in E^+} -\log \alpha_{i,j} + \lambda \sum_{(i,j) \in E^-} -\log(1 - \alpha_{i,j})}_{\text{For GAT Moudle}},$$

where $E^+$ represents the set of nodes pairs that should be connected based on prior knowledge or inferred relationships, $E^-$ represents the set of pairs that "should not be connected," $\alpha_{i,j}$ is the weight calculated by GAT, and $\lambda$ is a balancing hyperparameter. This design allows GAT to automatically adjust weights using prior or detected relationships during the learning of frame-level attention allocation while suppressing interference from noisy edges.

Combining the above multiple objectives, the total loss function $\mathcal{L}_{total}$ can be expressed in Exp. (16):

$$\mathcal{L}_{total} = \alpha_{cls}\mathcal{L}_{cls} + \alpha_{ret}\mathcal{L}_{ret} + \alpha_{GAT}\mathcal{L}_{GAT}.$$

Where $\alpha_{cls}$, $\alpha_{ret}$ and $\alpha_{GAT}$ are weighting coefficients for the three loss components, used to adjust the relative importance of different task objectives during training. By performing end-to-end backpropagation with this total loss function, this method not only learn the discriminative capabilities required by the classifier and retriever for cross-modal matching but also refines and enhances the attention weights in the GAT module, ultimately achieving more comprehensive spatiotemporal relationship modeling and semantic understanding.

## 5. Experiment

In this section, relevant experimental performance and analysis are presented.

### 5.1. Dataset and Evaluation Metrics

For the detection task, the experiment employed two widely recognized datasets: UCF-Crime [18] and XD-Violence [19]. XD-Violence, the largest publicly available dataset for violence monitoring, includes 4,754 videos totaling 217 hours and covers six categories of violent incidents: verbal abuse, car accidents, explosions, fights, riots, and shootings. The dataset is divided into a training set of 3,954 videos and a test set of 800 videos, with the latter comprising 500 violent and 300 non-violent videos. The UCF-Crime dataset consists of 1,900 real-world surveillance videos, with 1,610 allocated for training and 290 for testing.

Regarding evaluation metrics, the experiment adheres to the framework described in Section 3.1. For XD-Violence, the metric $\mathcal{AP}^\mathcal{M}$ from Exp. (1) is used, while for UCF-Crime, the metric $\mathcal{A}_\mathcal{M}$ from Exp. (2) is applied. This approach aligns with prior research [1][5][7] [13]-[28].

For the retrieval task, the experiment utilized the UCFCrime-AR [6] dataset, which contains 1,900 real-world surveillance videos accompanied by corresponding bilingual text descriptions. Of these, 1,610 videos are used for training and 290 for testing. Evaluation follows the setup outlined in Section 3.2. Specifically, the retrieval metric R@k from Exp. (3) is applied to the XD-Violence dataset, consistent with methodologies from earlier studies [6][31]-[35].

### 5.2. Experimental Settings

In the configuration of some pretraining modules, the multimodal model used is ImageBind [13], with the version imagebind_huge. For knowledge graph construction, the object detection model for Entity 2 is YOLO-World [15], specifically the YOLO-Worldv2-X. The software utilized for KG storage is ArangoDB. In the configuration of the large language model (LLM), the Glaude 3.5 Sonnet API is invoked.

For hyperparameter settings, In Exp. (6) and Exp. (8), $\sigma$ is set to 0.25 and 3, respectively. In Exp. (14), $\alpha$ is set to 0.9, In Exp. (15), $\lambda$ is set to 1 and in Exp. (16), $\alpha_{cls}$, $\alpha_{ret}$ and $\alpha_{GAT}$ are set to 1.4, 1.3, and 1, respectively.

The training hardware consists of NVIDIA Tesla A100 40GB, and the system employs the Adam optimizer during training. The initial learning rate is set to $5 \times 10^{-5}$, with a multiplicative decay factor of 0.95 per epoch.

### 5.3. Comparison with State-of-the-art Methods

Table 1 aims to compare the proposed TIO system with previous studies on fine-grained violence detection in the XD-Violence datasets. As shown in Table 1, the experimental results of the proposed TIO system on the XD-Violence dataset demonstrate significant performance advantages, particularly in its detection capability in highly dynamic and subtle anomaly scenarios.

In the Abuse category, the proposed TIO system achieves an $\mathcal{AP}^\mathcal{M}$ of 17.84, which is a 6.86% improvement compared to the best baseline, MISSIONGNN (10.98), showcasing its outstanding ability in low-dynamic feature scenarios. In the Car Accident category, the proposed TIO system achieves an $\mathcal{AP}^\mathcal{M}$ of 28.89, outperforming RTFM (25.36) and S3R (23.82), indicating its effectiveness in capturing subtle anomalies in car accident scenes.

In the more dynamic Explosion and Fighting categories, the proposed TIO system achieves $\mathcal{AP}^\mathcal{M}$ of 73.54 and 79.98, respectively. In the Shooting category, the proposed TIO system achieves an $\mathcal{AP}^\mathcal{M}$ of 32.65, significantly surpassing AnomalyCLIP (26.13) by 6.52%, demonstrating its adaptability to high-speed dynamic and explosive sound scenarios.

These results clearly demonstrate that the proposed TIO system leverages a combination of KG and GAT modules to effectively enhance anomaly detection performance. The KG helps model semantic relationships between different anomalous behaviors, while GAT further strengthens the selective attention to anomalous features. Moreover, the dual-branch design of the proposed TIO system achieves higher performance compared to single-branch systems designed specifically for violence detection. This advantage is due to the dual-branch architecture's ability to separately focus on visual and auditory feature modeling, enabling effective multimodal information fusion.

**Table 1**

Fine-Grained Result of the satate-of-the-art (SoTA) methods and the proposed TIO on XD-Violence ($\mathcal{AP}^{\mathcal{M}}$)

| Method | Class | | | | | |
|---|---|---|---|---|---|---|
| | Abuse | CarAccident | Explosion | Fighting | Riot | Shooting |
| CLIP [26] | 0.32 | 12.21 | 22.26 | 25.25 | 66.60 | 1.26 |
| RTFM [23] | 9.25 | 25.36 | 53.53 | 61.73 | 90.38 | 18.01 |
| S3R [24] | 2.63 | 23.82 | 45.29 | 49.88 | 90.41 | 4.34 |
| ActionCLIP [27] | 2.73 | 25.15 | 55.28 | 58.09 | 89.31 | 12.87 |
| AnomalyCLIP [28] | 6.10 | **31.31** | 68.75 | 71.44 | **92.74** | 26.13 |
| MISSIONGNN [30] | 10.98 | 8.91 | 42.49 | 34.06 | 81.64 | 15.03 |
| TIO (Ours) | **17.84** | 28.89 | **73.54** | **79.98** | 82.64 | **32.65** |

**Table 2**

Fine-Grained Result of the SoTA methods and the proposed TIO on UCF-Crime ($\mathcal{A}_{\mathcal{M}}$)

| Method | Class | | | | | | | | | | | | |
|---|---|---|---|---|---|---|---|---|---|---|---|---|---|
| | Abuse | Arrest | Arson | Assault | Burglary | Explosion | Fighting | RoadAcc | Robbery | Shooting | Shoplifting | Stealing | Vandalism |
| CLIP [26] | 57.37 | 80.65 | 91.72 | 80.83 | 74.34 | 90.31 | 83.54 | 87.46 | 70.22 | 63.99 | 71.21 | 45.49 | 66.45 |
| RTFM [23] | 79.99 | 62.57 | 90.53 | 82.27 | 85.53 | 92.76 | 85.21 | 90.31 | 81.17 | 82.82 | 92.56 | 90.23 | 87.20 |
| S3R [24] | 86.38 | 68.45 | 92.19 | 93.55 | 86.91 | 93.55 | 81.69 | 85.03 | 82.07 | 85.32 | 91.64 | 94.59 | 83.82 |
| STRL [25] | 95.33 | 79.26 | 93.27 | 91.74 | 89.06 | 92.25 | 87.36 | 80.24 | 87.75 | 84.5 | 92.31 | 94.22 | 88.19 |
| ActionCLIP [27] | 91.88 | 90.47 | 89.21 | 86.87 | 81.31 | 94.08 | 83.23 | 94.34 | 82.82 | 70.53 | 91.60 | 94.06 | 89.89 |
| AnomalyCLIP [28] | 75.03 | **94.56** | **96.66** | 94.80 | 90.08 | 94.79 | 88.76 | 93.30 | 86.85 | 87.45 | 89.47 | 97.00 | 89.78 |
| MISSIONGNN [30] | 93.68 | 92.95 | 94.15 | 96.89 | 92.31 | 94.79 | 92.38 | 97.16 | 81.04 | 90.57 | 92.94 | 97.74 | 84.03 |
| TIO (Ours) | **96.32** | 93.68 | 96.08 | **97.02** | **93.55** | **94.88** | **93.23** | **97.66** | **87.78** | **92.56** | **93.32** | **97.81** | **91.62** |

Similarly, as shown in Table 2, the proposed TIO system demonstrates outstanding performance across various categories in the UCF-Crime dataset, significantly outperforming competing methods in most scenarios. Specifically, TIO achieves the highest $\mathcal{A}_{\mathcal{M}}$ in key categories such as Abuse (96.32), Assault (97.02), RoadAcc (97.66), Stealing (97.81), and Vandalism (91.62). Compared to its closest competitor, MISSIONGNN, the proposed TIO exhibits notable improvements across multiple categories, particularly excelling in scenarios requiring fine-grained anomaly detection. For instance, in the RoadAcc and Stealing categories, the prosed TIO's $\mathcal{A}_{\mathcal{M}}$ scores are significantly higher than those of MISSIONGNN and AnomalyCLIP, showcasing its superior ability to detect subtle and complex anomalies in dynamic scenarios.

In high-dynamic categories such as Explosion (94.88), Fighting (93.23), and Shooting (92.56), the proposed TIO significantly outperforms AnomalyCLIP and other baseline methods, demonstrating its robust modeling capability of audiovisual features in high-energy environments. Notably, in the Shooting category, the proposed TIO achieves an $\mathcal{A}_{\mathcal{M}}$ of 92.56, which is considerably higher than AnomalyCLIP's 87.45, highlighting its precise capability in capturing rapid motion and auditory signals. Similarly, in low-dynamic categories such as Abuse and Stealing, the proposed TIO's dual-branch design delivers excellent performance, fully leveraging the advantages of multimodal information fusion to achieve precise detection of subtle anomalies.

These results further underscore the architectural advantages of TIO, particularly its dual-branch design, which independently models visual and auditory features while effectively integrating multimodal information. This enables TIO to achieve consistently high AUC scores across diverse scenarios, significantly outperforming single-branch systems such as RTFM, S3R, and STRL. Furthermore, the proposed TIO incorporates KG and GAT modules, which allow it to model semantic relationships between anomalous behaviors and selectively enhance attention to critical features. This further solidifies its outstanding performance across all categories.

**Table 3**

Course-Grained Result of the SoTA methods and the proposed TIO on XD-Violence ($\mathcal{AP}^{\mathcal{M}}$) and UCF-Crime ($\mathcal{A}_{\mathcal{M}}$)

| Method | XD-Violence $\mathcal{AP}^{\mathcal{M}}$ | UCF-Crime $\mathcal{A}_{\mathcal{M}}$ |
|---|---|---|
| CLIP [26] | 27.21 | 58.63 |
| RTFM [23] | 77.81 | 84.03 |
| S3R [24] | 80.26 | 85.99 |
| STRL [25] | (~) | 87.43 |
| ActionCLIP [27] | 61.01 | 82.30 |
| AnomalyCLIP [28] | 78.51 | 86.36 |
| MISSIONGNN [30] | 98.42 | 84.48 |
| TIO (Ours) | **99.25** | **88.67** |

Table 3 presents the coarse-grained results of the proposed TIO system on the XD-Violence and UCF-Crime datasets, both achieving the best performance and demonstrating significant advantages. On the XD-Violence dataset, the proposed TIO achieved an $\mathcal{AP}^{\mathcal{M}}$ of 99.25, which is nearly perfect and significantly outperforms other methods, particularly surpassing the second-best MISSIONGNN, which achieved 98.42 $\mathcal{AP}^{\mathcal{M}}$ score. On the UCF-Crime dataset, the $\mathcal{A}_{\mathcal{M}}$ value of the proposed TIO system is 88.67, once again setting a new benchmark among state-of-the-art methods. It exceeds the second-best STRL (87.43) and other competing methods, such as AnomalyCLIP (86.36) and S3R (85.99), demonstrating superior stability and fine-grained anomaly detection capability. Particularly in tasks requiring simultaneous handling of both high-dynamic and low-dynamic scenarios, the proposed TIO effectively balances multimodal features to achieve stable and superior performance.

In contrast, traditional methods such as CLIP (58.63) and ActionCLIP (82.30) perform relatively weaker, indicating certain limitations in adapting to scenarios involving multimodal fusion and anomaly detection. Further analysis shows that although MISSIONGNN comes close to the proposed TIO in performance on the XD-Violence dataset, its $\mathcal{A}_\mathcal{M}$ score on the UCF-Crime dataset (84.48) is significantly lower, indicating it is less adaptive to diverse scenarios compared to the proposed TIO. Additionally, while STRL demonstrates stable performance on UCF-Crime, its lack of results on XD-Violence limits its comprehensiveness.

The proposed TIO system leverages an innovative dual-branch architecture that independently models visual and auditory features, aided by KG to further explore semantic relationships between anomalous behaviors. This approach effectively enhances attention to critical features, enabling precise anomaly detection in diverse scenarios. Experimental results demonstrate that the proposed TIO system excels in handling high-dynamic anomaly detection tasks by utilizing its dual-branch design, multimodal information fusion, and integration of GAT with KG. Consequently, the proposed TIO achieves significant performance improvements across diverse scenarios in both XD-Violence and UCF-Crime datasets.

**Table 4**

Comparisons with the SOTA methods on UCFCrime-AR (R@K)

| Method | Video→Text | | |
|---|---|---|---|
| | R@1 | R@5 | R@10 |
| Random Baseline | 0.3 | 1.0 | 3.1 |
| CE [31] | 5.5 | 19.7 | 32.4 |
| MMT [32] | 7.2 | 23.1 | 39.0 |
| T2VLAD [33] | 6.2 | 27.9 | 43.1 |
| X-CLIP [34] | 6.9 | 25.8 | 40.3 |
| HL-Net [11] | 5.5 | 22.8 | 35.5 |
| XML [35] | 6.6 | 25.9 | 43.4 |
| ALAN [6] | 7.3 | 24.8 | 46.9 |
| TIO(Ours) | **9.8** | **32.1** | **68.8** |

Table 4 presents the ranking results (R@K) of the proposed TIO system and existing SoTA methods on the UCFCrime-AR dataset for the Video→Text task. The results indicate that the proposed TIO system achieved the best performance across all ranking metrics (R@1, R@5, and R@10), demonstrating its exceptional retrieval capabilities. For the R@1 metric, the proposed TIO achieved a score of 9.8, significantly surpassing the ALAN, which scored 7.3. Compared to traditional methods such as CE and HL-Net (both scoring 5.5 for R@1), the advantage of TIO is even more pronounced. For R@5 and R@10, the proposed TIO achieved scores of 32.1 and 68.8, respectively, outperforming ALAN (24.8 and 46.9) by 7.3% and 21.9%. This demonstrates the proposed TIO's stable advantage in both medium and large-scale retrieval ranges. Compared to other methods like T2VLAD (R@10 = 43.1) and XML (R@10 = 43.4), the proposed TIO achieved over a 25% improvement in R@10, highlighting its strong adaptability in large-scale retrieval scenarios. Further analysis shows that the proposed TIO consistently outperformed the random baseline across all ranking metrics. For instance, the random baseline achieved only 3.1% for R@10, underscoring that the proposed TIO's superior performance is a direct result of its innovative architectural design. Compared to the ALAN, while ALAN demonstrated some competitiveness in limited retrieval ranges (e.g., R@1 and R@5), its performance declined significantly for larger retrieval ranges (e.g., R@10). In contrast, the proposed TIO maintained stable and significantly improved performance, reflecting its robustness in handling large-scale retrieval tasks. Traditional methods such as CE and HL-Net performed relatively poorly across all metrics, with CE achieving only 32.4% for R@10—less than half of the proposed TIO's score—highlighting the limitations of these methods in multimodal feature modeling and scenario adaptability.

In conclusion, the proposed TIO system leverages efficient feature extraction through KG and GAT, as well as its multi-branch architecture design, to achieve outstanding performance in the Video→Text task on the UCFCrime-AR dataset. Moreover, experimental results demonstrate that the prosed TIO excels not only in anomaly detection tasks but also in cross-modal retrieval scenarios, showcasing its strong adaptability and stability.

**5.4. Ablation Study**

Table 5 presents the ablation study results on the XD-Violence dataset and the UCF-Crime dataset, exploring the impact of the KG, GAT, and Time Temporal Sequence Module on the performance of the proposed TIO system.

**Table 5**

Ablation Study in XD-Violence ($\mathcal{AP}^\mathcal{M}$) and UCF-Crime ($\mathcal{A}_\mathcal{M}$)

| KG | GAT | Time Temporal Sequence Module | $\mathcal{AP}^\mathcal{M}$ | $\mathcal{A}_\mathcal{M}$ |
|---|---|---|---|---|
| ✔ | | | 78.12 | 66.54 |
| | ✔ | | 81.45 | 70.32 |
| | | ✔ | 76.78 | 64.89 |
| ✔ | ✔ | | 87.34 | 76.12 |
| | ✔ | ✔ | 84.56 | 72.89 |
| ✔ | | ✔ | 85.21 | 74.23 |
| ✔ | ✔ | ✔ | 99.25 | 88.67 |

The results indicate that enabling a single module improves the system's performance, though the effect is limited. For instance, with only KG enabled, the $\mathcal{AP}^\mathcal{M}$ on XD-Violence is 78.12, and the $\mathcal{A}_\mathcal{M}$ on UCF-Crime is 66.54, demonstrating that KG enhances the system's understanding of anomalous behaviors but is insufficient on its own for achieving high performance. When only GAT is enabled, the $\mathcal{AP}^\mathcal{M}$ improves to 81.45 on XD-Violence, and the $\mathcal{A}_\mathcal{M}$ increases to 70.32 on UCF-Crime, suggesting that GAT effectively models semantic relationships between anomalous features. Enabling only the Time Temporal Sequence Module results in an $\mathcal{AP}^\mathcal{M}$ of 76.78 on XD-Violence and an $\mathcal{A}_\mathcal{M}$ of 64.89 on UCF-Crime, showing that its contribution is relatively smaller and requires integration with other modules to maximize its potential. When two modules are combined, the system's performance improves significantly. For example, combining KG and GAT yields an $\mathcal{AP}^\mathcal{M}$ of 87.34 on XD-Violence and an $\mathcal{A}_\mathcal{M}$ of 76.12 on UCF-Crime, highlighting their complementary roles in capturing multimodal features and semantic relationships. The combination of GAT and the Time Temporal Sequence Module achieves an $\mathcal{AP}^{\mathcal{M}}$ of 84.56 on XD-Violence and an $\mathcal{A}_{\mathcal{M}}$ of 72.89 on UCF-Crime, indicating that GAT supports temporal modeling but is slightly less effective than the KG and GAT combination. The KG and Time Temporal Sequence Module combination yields an $\mathcal{AP}^{\mathcal{M}}$ of 85.21 on XD-Violence and an $\mathcal{A}_{\mathcal{M}}$ of 74.23 on UCF-Crime, showing that KG provides strong contextual support for the temporal module. When all three modules are enabled simultaneously, the system achieves optimal performance, with an $\mathcal{AP}^{\mathcal{M}}$ of 99.25 on XD-Violence and an $\mathcal{A}_{\mathcal{M}}$ of 88.67 on UCF-Crime. This demonstrates that KG provides semantic contextual information, GAT enhances selective attention to anomalous features, and the Time Temporal Sequence Module captures the temporal dynamics of anomalous behaviors, creating maximum synergy in multimodal modeling. These results highlight that while individual modules improve performance, the combination of KG, GAT, and the Time Temporal Sequence Module is key to achieving the best anomaly detection performance. Their synergistic effect significantly enhances the adaptability and detection capabilities of the proposed TIO system in diverse scenarios.

**Table 6**

Ablation Study in UCFCrime-AR (R@K)

| KG | GAT | Time Temporal Sequence Module | R@1 | R@5 | R@10 |
|---|---|---|---|---|---|
| ✔ | | | 5.4 | 21.8 | 52.7 |
| | ✔ | | 6.2 | 23.4 | 56.1 |
| | | ✔ | 4.9 | 20.1 | 49.8 |
| ✔ | ✔ | | 8.1 | 28.3 | 62.5 |
| | ✔ | ✔ | 7.4 | 26.8 | 60.2 |
| ✔ | | ✔ | 7.9 | 27.5 | 61.3 |
| ✔ | ✔ | ✔ | **9.8** | **32.14** | **68.8** |

Table 6 shows the ablation study in UCFCrime-AR, the experimental results demonstrate that the Knowledge Graph (KG) and Graph Attention Network (GAT) play a critical role in retrieval tasks. When only KG is enabled, retrieval performance (R@K) improves significantly, indicating that the knowledge graph effectively structures retrieval cues and enhances semantic relationships. Similarly, when only GAT is enabled, the performance also improves markedly, showing that GAT can effectively capture relationships between key nodes and strengthen feature attention.

When KG and GAT are combined (KG + GAT), the system performance further improves, demonstrating the synergistic effect between these two modules. Particularly for R@1 and R@5, the combination results in substantial performance gains, suggesting that this integration better balances semantic modeling and feature attention, thereby significantly enhancing retrieval accuracy.

Additionally, with the inclusion of the Time Temporal Sequence Module, the system considers temporal relationships, supplementing contextual information in the retrieval process, which leads to a more outstanding performance in R@10. When all modules are fully integrated, the proposed TIO system achieves optimal performance (R@1 = 9.8, R@5 = 32.14, R@10 = 68.8).

In summary, the combination of the KG and GAT is the core driving factor behind the improvement in retrieval performance. The addition of the Time Temporal Sequence Module further enhances the system's temporal awareness, enabling the overall retrieval performance to reach its optimal state.

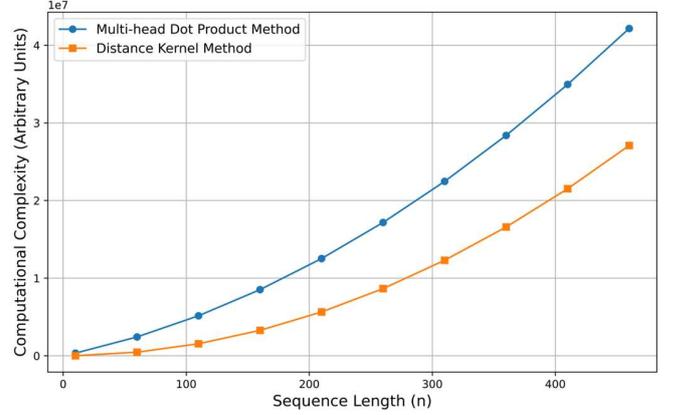

Fig.6. Computational Complexity for proof 1.

Figure 6 illustrates the variation in computational complexity of the Multi-head Dot Product Method and the Distance Kernel Method as the sequence length $n$ increases. It can be observed that while the computational complexity of both methods grows significantly with increasing sequence length $n$, their growth rates differ notably. For the Multi-head Dot Product Method, its computational complexity curve is steeper, primarily due to the quadratic complexity of attention computation $T_{atten} = O(Hn^2d)$, which dominates as $n$ increases. Although this method also includes the linear projection cost $T_{proj} = O(HnDd)$, its contribution is relatively minor compared to the quadratic attention computation. As a result, the total computational cost of this method increases sharply with larger sequence lengths. In contrast, the Distance Kernel Method also exhibits quadratic growth in computational complexity, dominated by $T_{dist-calc} = O(n^2D)$, which arises from the cost of pairwise Euclidean distance calculations. However, its growth rate is noticeably slower than that of the Multi-head Dot Product Method because it is unaffected by additional parameters such as the number of heads $H$ and the projection dimension $d$. In scenarios involving short sequences, the computational costs of the two methods are comparable, but as the sequence length grows, the cost of the Multi-head Dot Product Method quickly surpasses that of the Distance Kernel Method and becomes significantly higher for long sequences. This indicates that the Distance Kernel Method, with its lower growth rate, is more computationally efficient for large-scale data processing, while the Multi-head Dot Product Method may be more suitable for short-sequence scenarios, where its computational cost remains manageable, and its flexible attention mechanism provides additional benefits. In conclusion, the experimental results clearly demonstrate that for larger sequence lengths, the computational cost of the Distance Kernel Method is significantly lower than that of the Multi-head Dot Product Method. This aligns with theoretical complexity analysis and further validates the computational efficiency of the Distance Kernel Method for large-scale data processing.

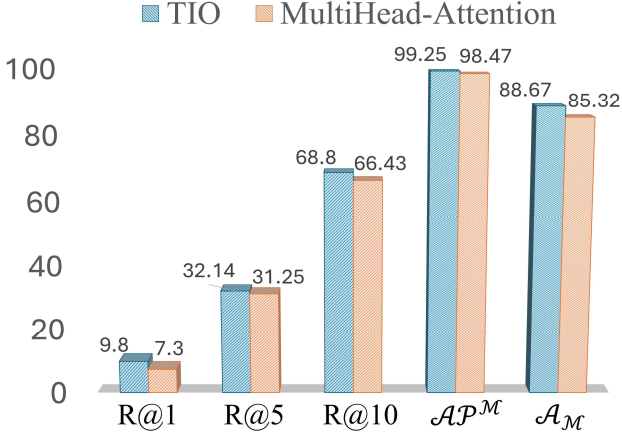

**Fig.7**. Compare with Multi-Head Attention

Figure 7 aims to show the ablation study for the proposed TIO system outperforms the Multi-Head Attention method across all retrieval metrics (R@1, R@5, R@10) and anomaly detection metrics $\mathcal{AP}^{\mathcal{M}}$ and $\mathcal{A}_{\mathcal{M}}$.

As show in Fig.7, In the retrieval task, the proposed TIO achieves a score of 9.8 for R@1, significantly higher than Multi-Head Attention's 7.3, demonstrating its superior capability in ensuring the relevance of high-precision retrieval results. For R@5 and R@10, the proposed TIO scores 32.14 and 68.8, respectively, also outperforming Multi-Head Attention's 31.25 and 66.43, showcasing its stability and accuracy in broader retrieval scenarios. In anomaly detection tasks, the proposed TIO achieves $\mathcal{AP}^{\mathcal{M}}$ =99.25 on XD-Violence and $\mathcal{A}_{\mathcal{M}}$ =88.67 on UCF-Crime, both surpassing Multi-Head Attention's 98.47 and 85.32, indicating the TIO's exceptional performance violence detection and violence retrieval.

The strength of the proposed TIO lies in its novel computation method for self-attention, which is based on the Distance Kernel Method and incorporates a temporal sequence module. Unlike existing approaches, the proposed TIO's primary distinction is its use of relative distances rather than feature similarity for self-attention calculation, making the model more lightweight. By focusing on computing the relative distances between frames and avoiding high-dimensional feature similarity computations, this method significantly reduces the complexity of attention computation. As a result, it lowers computational costs while substantially improving the efficiency of training and inference. The ablation study demonstrates that this combination of the Distance Kernel Method and the temporal sequence module enables TIO to achieve both performance and efficiency gains in retrieval and violence detection tasks.

**5.5. Visualization**

Figure 8 visualization illustrates the analysis of violent scenes, showcasing how the integration of KG and GAT significantly enhances the model's interpretability. The image captures and highlights multiple violent scenarios, including physical conflicts, pursuits, and other violent actions. The highlighted areas on the heatmap (in red and yellow) visually demonstrate the key regions the model focuses on when detecting violent behaviors. These highlighted regions are often concentrated on body movements, interactions between individuals, and relevant contextual details, indicating the model's capability to accurately identify features associated with violent activities.

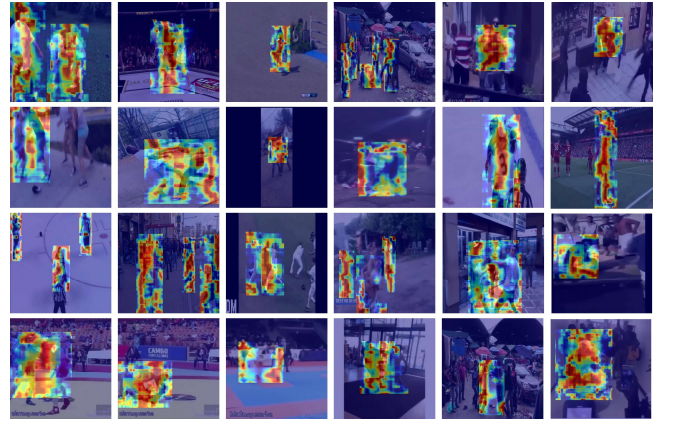

**Fig.8**. Heatmap Visualization

The KG plays a critical role in detecting violent scenes. By incorporating semantic background information, the knowledge graph models the relationships between individuals, objects, and scenes, linking them to potential violent behaviors. This semantic enhancement allows the model to understand the latent violent factors in a scene from a higher-level perspective and focus its attention on violence-related areas. Meanwhile, the GAT further strengthens this mechanism by capturing dynamic relationships between frame-level features. It effectively integrates contextual information across multiple frames, enabling the model to focus on the temporal continuity and spatial distribution characteristics of violent behaviors.

This visualization also reflects the proposed TIO system performance across diverse violent scenarios. For instance, in crowded environments, the TIO system accurately identifies the regions of physical collisions between individuals. In chase or confrontation scenarios, the proposed TIO's attention is directed towards fast-moving individuals or specific action details. Such performance demonstrates the broad applicability of the KG and GAT module in violent scene detection.

Figure 9 illustrates a simple example of traffic accident detection (RoadAcc). In the experiment, the SAM2 [37] model was used to precisely lock onto objects in the scene. As shown in the Fig.9, when the model's heatmap focuses on the two vehicles in the scene, the system can clearly identify the occurrence of a traffic accident. This process highlights the importance of KG and GAT modules. The KG helps the model structurally represent semantic relationships in the scene, such as the distance and interactions between vehicles, while GAT enhances the model's attention to critical features, enabling it to effectively capture anomalous events. The overall detection curve demonstrates that the decision score steadily increases as the scene gradually enters an abnormal state, exceeding the threshold within the detection region, and ultimately successfully identifies the anomalous situation. This approach, which integrates multimodal information and contextual modeling, is key to accurately detecting traffic accidents and other fine-grained anomalous behaviors.

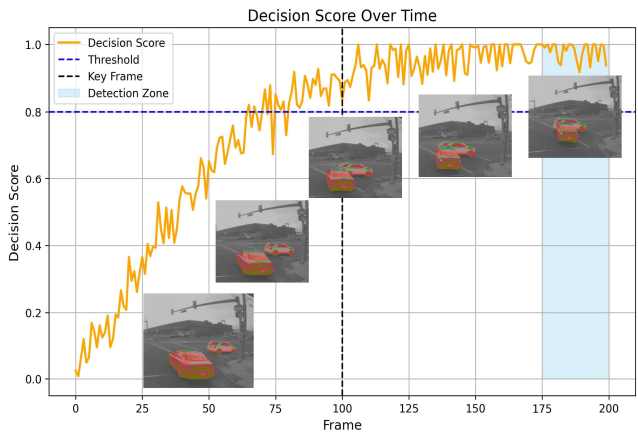

**Fig.9**. A Simple Example of RoadAcc

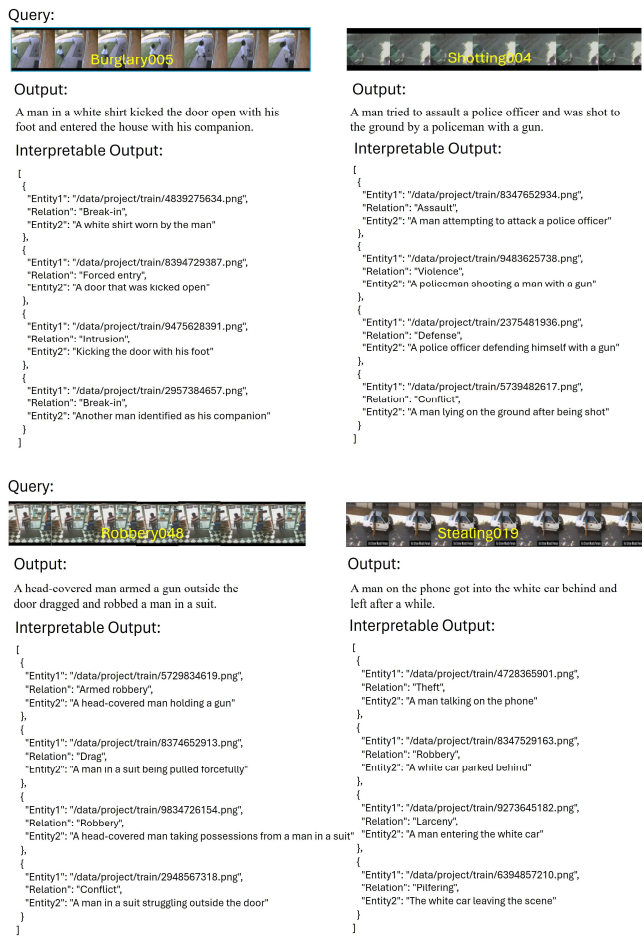

**Fig.10**. Simple Example of Violence Retrieval

Figure 10 illustrates examples of violence retrieval with explainability provided by the TIO system. For each query video, the system generates corresponding retrieval outputs and presents interpretable triples in natural language form, clearly describing the semantic relationships between events, actions, and objects within the scene. Through this approach, the TIO system not only retrieves videos related to the query but also provides explainable and structured information to help users understand the details and semantics of violent behaviors.

In the example "Burglary005," the TIO system successfully extracts key frames and represents the video content in the form of triples. For instance, it explicitly indicates that "A white shirt worn by the man" is associated with "Break-in," and "Kicking the door with his foot" is identified as a key action demonstrated in the video. Similarly, in the example "Shooting004," the system describes the violent behavior of "A policeman shooting a man with a gun" and explains the relationships between entities in the event using keywords such as "Violence" and "Defense." This explanatory representation enhances the semantic transparency of the retrieval results.

Furthermore, in the examples "Robbery048" and "Stealing019," the TIO system demonstrates its ability to provide explainability in complex scenarios. For the former, the system extracts fine-grained semantic information such as "A head-covered man holding a gun" and "A man in a suit struggling outside the door," associating them with keywords like "Armed robbery" and "Conflict" to accurately depict the key elements of the robbery scene. For the latter, the system uses keywords such as "Theft" and "Larceny" to explain the actions and relationships between "A man talking on the phone" and "A white car leaving the scene," clearly presenting the behavioral details.

These examples show that the TIO system not only possesses precise retrieval capabilities but also generates query-related triples to provide users with intuitive and semantically transparent results, significantly enhancing the model's explainability. This capability is of great practical significance for the monitoring, analysis, and enforcement of violent behaviors.

### 5.6. Case Study

Using the proposed TIO system, an experiment was conducted to analyze the impact of bystander presence on the outcomes of violent behaviors. The retrieval results revealed an interesting phenomenon: in both the fighting and non-violent categories, the number of bystanders significantly influenced the outcomes of violent incidents. When more bystanders were present, conflicts often remained at the verbal level, manifesting as insults, arguments, and other forms of verbal conflict. For instance, one extracted triple relation described a scene where "a man shouted at another person in a crowd," annotated as "Verbal conflict," with accompanying information about the number of bystanders—"approximately 10 onlookers." Additionally, the system retrieved behaviors such as "bystanders attempting to separate the conflicting parties," clearly illustrating the role of bystander intervention in de-escalating conflicts. This phenomenon suggests that the presence of bystanders can, to some extent, reduce the risk of violence escalation through deterrence or direct intervention.

In contrast, when fewer bystanders were present, the nature of the conflict underwent a significant shift. Analysis of the extracted triple relations revealed that conflicts were more likely to escalate from verbal to physical violence. For example, one triple described a scene where "a man assaulted a passerby in an alley," annotated as "Physical conflict," with the detail that "no noticeable bystanders were present." This indicates that in the absence of bystander intervention, potential aggressors may feel less deterred or socially constrained, leading to an escalation into physical violence. Additionally, retrieved relations revealed that physical violence often stemmed from prior verbal disputes, such as "intense verbal arguments escalating into punching and kicking," further underscoring the correlation between bystander absence and violent escalation.

This KG-based analysis not only confirmed the impact of bystander presence on the nature of violent behavior but also demonstrated the proposed TIO system's ability to interpret complex scenarios through structured semantic information. The system extracts triple relations from videos and texts, presenting bystander data, conflict types, and contextual features in an intuitive manner, offering highly interpretable retrieval results. These findings highlight the potential role of bystanders in conflict dynamics, providing new directions for studying the mechanisms of social behavior and practical guidance for monitoring and intervention strategies in violent situations.

## 6. Conclusions

This study presents an innovative Three-in-One (TIO) system that provides a solution for weakly supervised violence detection (WSVD) and violence retrieval (VAR) tasks, significantly enhancing interpretability, system performance, and user experience. By introducing KG, GAT module and a dual-branch design, the system achieves end-to-end multi-task processing. It not only addresses the interpretability limitations caused by the black-box nature of models in current research but also optimizes complexity through the use of kernel function-based Euclidean distance in unified multimodal embeddings and graph attention networks. The lightweight design of the temporal module further improves computational efficiency and model performance.

In addition, compared to traditional independent model designs, the TIO system employs a unified multi-task model to reduce redundancy in design and training processes, promote information sharing and collaborative usage, and optimize semantic reasoning capabilities, thereby achieving dual improvements in system efficiency and user experience. Experimental results demonstrate that the proposed TIO system achieves outstanding performance on benchmark datasets such as XD-Violence, UCF-Crime and UCFCrime-AR, highlighting its extensive potential and research value in the fields of violence detection and retrieval. Furthermore, case studies validate its significance in sociology and the study of human activities.

## Authors' Biographies

**Wen-Dong Jiang** received his Bachelor's degree in Multimedia and Game Science Development from Lunghwa University of Science and Technology, Taoyuan, Taiwan, in 2021. In 2022, he studied Software Engineering at Fayette Institute of Technology, Oak Hill, West Virginia, USA. He obtained his master's degree in computer science and information engineering from Ming Chuan University, Taoyuan, Taiwan, in 2023. Currently, he is pursuing a Ph.D. in Computer Science and Information Engineering at Tamkang University, Taiwan. His research interests include interpretable machine learning and AIoT.

**Chih-Yung Chang** received the Ph.D. degree in computer science and information engineering from the National Central University, Zhongli, Taiwan, in 1995. He is currently a Full Professor with the Department of Computer Science and Information Engineering, Tamkang University, New Taipei City, Taiwan. His current research interests include internet of things, wireless sensor networks, ad hoc wireless networks, and Long Term Evolution (LTE) broadband technologies. He has served as an Associate Guest Editor for several SCI-indexed journals, including the International Journal of Ad Hoc and Ubiquitous Computing from 2011 to 2024, the International Journal of Distributed Sensor Networks from 2012 to 2014, IET Communications in 2011, Telecommunication Systems in 2010, the Journal of Information Science and Engineering in 2008, and the Journal of Internet Technology from 2004 to 2018.

**Diptendu Sinha Roy** received the Ph.D. degree in engineering from the Birla Institute of Technology, Mesra, India, in 2010. In 2016, he joined the Department of Computer Science and Engineering, National Institute of Technology (NIT) Meghalaya, Shillong, India, as an Associate Professor, where he has been working as the Chair of the Department of Computer Science and Engineering since January 2017. Prior to his stint at NIT Meghalaya, he worked with the Department of Computer Science and Engineering, National Institute of Science and Technology, Berhampur, India. His current research interests include software reliability, distributed and cloud computing, and the Internet of Things (IoT), specifically applications of artificial intelligence/machine learning for smart integrated systems. Dr. Roy is a member of the IEEE Computer Society.

`